\documentclass[11pt,a4paper]{article}
\usepackage[hyperref]{acl2020}
\usepackage{times}
\usepackage{latexsym}
\usepackage{amsmath}
\usepackage{amssymb}
\usepackage{caption}
\usepackage{float}
\usepackage{multirow}
\usepackage{graphicx}
\usepackage{booktabs} 

\usepackage{microtype}

\aclfinalcopy 
\def\aclpaperid{23} 

\usepackage{svg}

\usepackage{xr}

\title{A Data-driven Approach for Noise Reduction in Distantly Supervised Biomedical Relation Extraction}

\author{
  Saadullah Amin\thanks{\quad Equal contribution\newline On behalf of the PRECISE4Q consortium}\quad Katherine Ann Dunfield\footnotemark[1]\quad Anna Vechkaeva\quad G\"unter Neumann\\\\German Research Center for Artificial Intelligence (DFKI)\\Multilinguality and Language Technology Lab\\\small\{\texttt{saadullah.amin, katherine.dunfield, anna.vechkaeva, guenter.neumann}\}\texttt{@dfki.de}
}

\date{}

\begin{document}
\maketitle
\begin{abstract}
Fact triples are a common form of structured knowledge used within the biomedical domain. As the amount of unstructured scientific texts continues to grow, manual annotation of these texts for the task of relation extraction becomes increasingly expensive. Distant supervision offers a viable approach to combat this by quickly producing large amounts of labeled, but considerably noisy, data. We aim to reduce such noise by extending an entity-enriched relation classification BERT model to the problem of multiple instance learning, and defining a simple data encoding scheme that \textit{significantly} reduces noise, reaching state-of-the-art performance for distantly-supervised biomedical relation extraction. Our approach further encodes knowledge about the direction of relation triples, allowing for increased focus on relation learning by reducing noise and alleviating the need for joint learning with knowledge graph completion.
\end{abstract}

\section{Introduction}

\begin{figure}[!t]
    \includegraphics[width=1.0\linewidth]{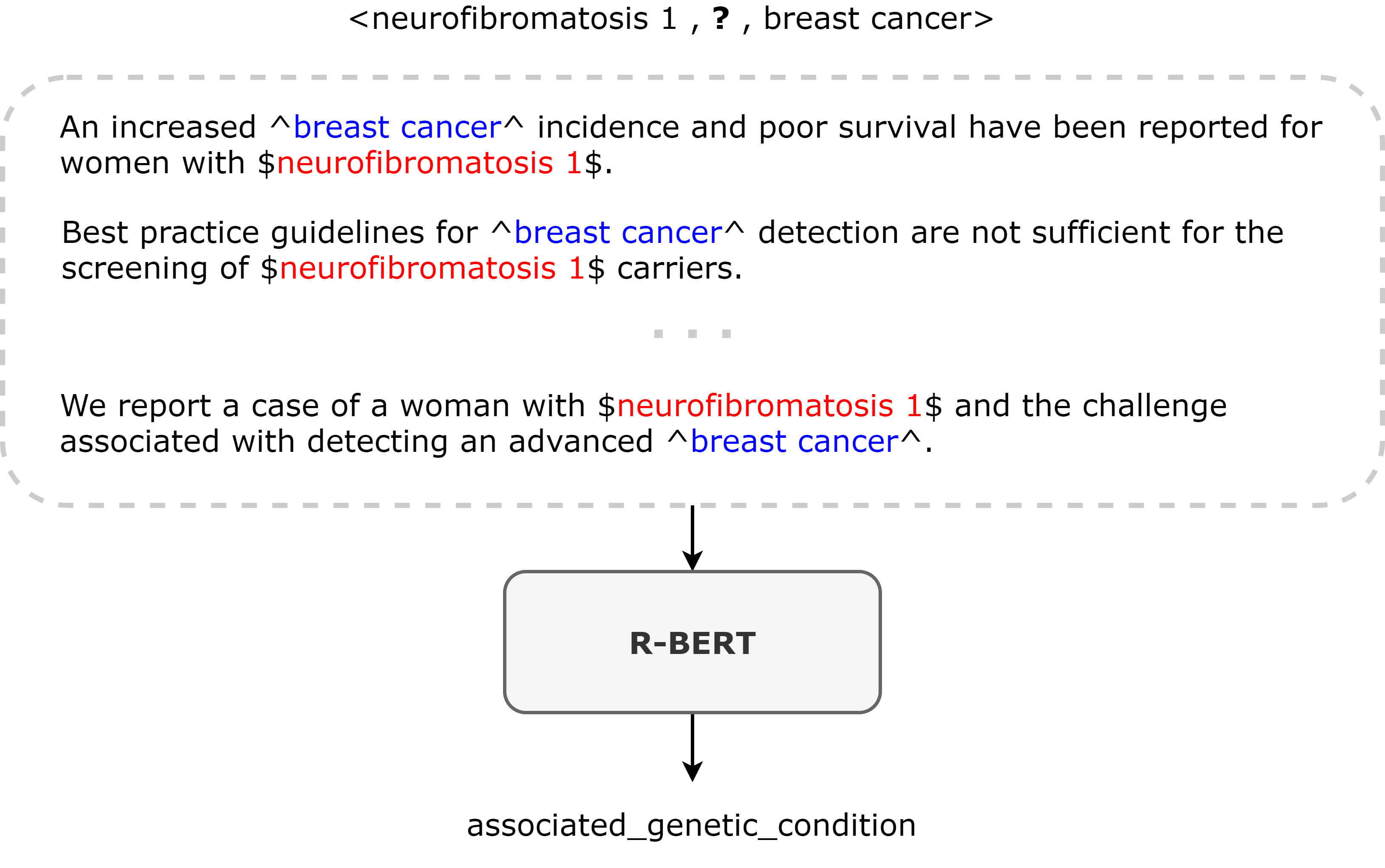}
    \centering
    \vspace{-0.5cm}
    \caption{Example of a distantly supervised bag of sentences for a knowledge base tuple (\textit{neurofibromatosis 1}, \textit{breast cancer}) with special order sensitive entity markers to capture the position and the latent relation direction with BERT for predicting the missing relation.}
    \label{fig:overview}
    \vspace{-0.2cm}
\end{figure}

Relation extraction (RE) remains an important natural language processing task for understanding the interaction between entities that appear in texts. In supervised settings \citep{guodong2005exploring,zeng2014relation,wang2016relation}, obtaining fine-grained relations for the biomedical domain is challenging due to not only the annotation costs, but the added requirement of domain expertise. Distant supervision (DS), however, provides a meaningful way to obtain large-scale data for RE \citep{mintz2009distant, hoffmann2011knowledge}, but this form of data collection also tends to result in an increased amount of noise, as the target relation may not always be expressed \citep{takamatsu2012reducing,ritter2013modeling}. Exemplified in Figure \ref{fig:overview}, the last two sentences can be seen as potentially \textit{noisy} evidence, as they do not explicitly express the given relation. 

Since individual instance labels may be unknown \citep{wang2018revisiting}, we instead build on the recent findings of \citet{wu2019enriching} and \citet{soares2019matching} in using positional markings and latent relation direction (Figure \ref{fig:overview}), as a signal to mitigate noise in \textit{bag-level} multiple instance learning (MIL) for distantly supervised biomedical RE. Our approach greatly simplifies previous work by \citet{dai2019distantly} with following contributions:
\begin{itemize}
    \itemsep0em 
    \item We extend \textit{sentence-level} relation enriched BERT \citep{wu2019enriching} to \textit{bag-level} MIL.
    \item We demonstrate that the simple applications of this model under-perform and require knowledge base order-sensitive markings, \textit{k-tag}, to achieve state-of-the-art performance. This data encoding scheme captures the latent relation direction and provides a simple way to reduce noise in distant supervision.  
    \item We make our code and data creation pipeline publicly available: \url{https://github.com/suamin/umls-medline-distant-re}
\end{itemize} 

\section{Related Work}

In MIL-based distant supervision for \textit{corpus-level} RE, earlier works rely on the assumption that at least one of the evidence samples represent the target relation in a triple \citep{riedel2010modeling,hoffmann2011knowledge,surdeanu2012multi}. Recently, piecewise convolutional neural networks (PCNN) \citep{zeng2014relation} have been applied to DS \citep{zeng2015distant}, with notable extensions in selective attention \citep{lin2016neural} and the modelling of noise dynamics \citep{luo2017learning}. \citet{han2018neural} proposed a joint learning framework for knowledge graph completion (KGC) and RE with mutual attention, showing that DS improves downstream KGC performance, while KGC acts as an indirect signal to filter textual noise. \citet{dai2019distantly} extended this framework to biomedical RE, using improved KGC models, ComplEx \citep{trouillon2017knowledge} and SimplE \citep{kazemi2018simple}, as well as additional auxiliary tasks of entity-type classification and named entity recognition to mitigate noise.

Pre-trained language models, such as BERT \citep{devlin2019bert}, have been shown to improve the downstream performance of many NLP tasks. Relevant to distant RE, \citet{alt2019fine} extended the OpenAI Generative Pre-trained Transformer (GPT) model \citep{radford2019language} for \textit{bag-level} MIL with selective attention \citep{lin2016neural}. \citet{sun2019ernie} enriched pre-training stage with KB entity information, resulting in improved performance. For \textit{sentence-level} RE, \citet{wu2019enriching} proposed an entity marking strategy for BERT (referred to here as R-BERT) to perform relation classification. Specifically, they mark the entity boundaries with special tokens following the order they appear in the sentence. Likewise, \citet{soares2019matching} studied several data encoding schemes and found marking entity boundaries important for \textit{sentence-level} RE. With such encoding, they further proposed a novel pre-training scheme for distributed relational learning, suited to few-shot relation classification \citep{han2018fewrel}.

Our work builds on these findings, in particular, we extend the BERT model \citep{devlin2019bert} for \textit{bag-level} MIL, similar to \citet{alt2019fine}. More importantly, noting the significance of \textit{sentence-ordered} entity marking in \textit{sentence-level} RE \citep{wu2019enriching,soares2019matching}, we introduce the \textit{knowledge-based} entity marking strategy suited to \textit{bag-level} DS. This naturally encodes the information stored in KB, reducing the inherent noise.

\section{Bag-level MIL for Distant RE}

\subsection{Problem Definition}

Let $\mathcal{E}$ and $\mathcal{R}$ represent the set of entities and relations from a knowledge base $\mathcal{KB}$, respectively. For $h, t \in \mathcal{E}$ and $r \in \mathcal{R}$, let $(h, r, t) \in \mathcal{KB}$ be a fact triple for an ordered tuple $(h, t)$. We denote all such $(h,t)$ tuples by a set $\mathcal{G}^{+}$, i.e., there exists some $r \in \mathcal{R}$ for which the triple $(h, r, t)$ belongs to the KB, called \textit{positive groups}. Similarly, we denote by $\mathcal{G}^{-}$ the set of \textit{negative groups}, i.e., for all $r \in \mathcal{R}$, the triple $(h, r, t)$ does not belong to KB. The union of these groups is represented by $\mathcal{G} = \mathcal{G}^{+} \cup \mathcal{G}^{-}$ \footnote{The sets are disjoint, $\mathcal{G}^{+} \cap \mathcal{G}^{-} = \varnothing$}. We denote by $\mathcal{B}_g = [s_g^{(1)}, ..., s_g^{(m)}]$ an unordered sequence of sentences, called \textit{bag}, for $g \in \mathcal{G}$ such that the sentences contain the group $g=(h, t)$, where the bag size $m$ can vary. Let $f$ be a function that maps each element in the bag to a low-dimensional relation representation $[\mathbf{r}_g^{(1)}, ..., \mathbf{r}_g^{(m)}]$. With $o$, we represent the bag aggregation function, that maps instance level relation representation to a final bag representation $\mathbf{b}_g=o(f(\mathcal{B}_g))$. The goal of distantly supervised \textit{bag-level} MIL for \textit{corpus-level} RE is then to predict the missing relation $r$ given the bag.

\label{sec:3.1}

\subsection{Entity Markers}

\citet{wu2019enriching} and \citet{soares2019matching} showed that using special markers for entities with BERT in the order they appear in a sentence encodes the positional information that improves the performance of \textit{sentence-level} RE. It allows the model to focus on target entities when, possibly, other entities are also present in the sentence, implicitly doing entity disambiguation and reducing noise. In contrast, for \textit{bag-level} distant supervision, the noisy channel be attributed to several factors for a given triple $(h, r, t)$ and bag $\mathcal{B}_g$:
\begin{enumerate}
    \itemsep0em 
    \item Evidence sentences may not express the relation.
    \item Multiple entities appearing in the sentence, requiring the model to disambiguate target entities among other.
    \item The direction of missing relation.
    \item Discrepancy between the order of the target entities in the sentence and knowledge base.
\end{enumerate}  
To address (1), common approaches are to learn a negative relation class \texttt{NA} and use better bag aggregation strategies \citep{lin2016neural, luo2017learning, alt2019fine}. For (2), encoding positional information is important, such as, in PCNN \citep{zeng2014relation}, that takes into account the relative positions of \textit{head} and \textit{tail} entities \citep{zeng2015distant}, and in \citep{wu2019enriching,soares2019matching} for \textit{sentence-level} RE. To account for (3) and (4), multi-task learning with KGC and mutual attention has proved effective \citep{han2018neural,dai2019distantly}. Simply extending sentence sensitive marking to \textit{bag-level} can be adverse, as it enhances (4) and even if the composition is uniform, it distributes the evidence sentence across several bags. On the other hand, expanding relations to multiple sub-classes based on direction \citep{wu2019enriching}, enhances class imbalance and also distributes supporting sentences. To jointly address (2), (3) and (4), we introduce KB sensitive encoding suitable for \textit{bag-level} distant RE.

Formally, for a group $g=(h,t)$ and a matching sentence $s_g^{(i)}$ with tokens $(x_0, ..., x_L)$\footnote{$x_0=$\texttt{[CLS]} and $x_L=$\texttt{[SEP]}}, we add special tokens \$ and \string^ to mark the entity spans as:

\noindent \textbf{Sentence ordered}: Called \textit{s-tag}, entities are marked in the order they appear in the sentence. Following \citet{soares2019matching}, let $s_1=(i, j)$ and $s_2=(k, l)$ be the index pairs with $0 <i < j-1, j < k, k \leq l-1$ and $l \leq L$ delimiting the entity mentions $e_1=(x_i, ..., x_j)$ and $e_2=(x_k, ..., x_l)$ respectively. We mark the boundary of $s_1$ with \$ and $s_2$ with \string^. Note, $e_1$ and $e_2$ can be either $h$ or $t$.

\noindent \textbf{KB ordered}: Called \textit{k-tag}, entities are marked in the order they appear in the KB. Let $s_h=(i, j)$ and $s_t=(k, l)$ be the index pairs delimiting head ($h$) and tail ($t$) entities, irrespective of the order they appear in the sentence. We mark the boundary of $s_h$ with \$ and $s_t$ with \string^.

The \textit{s-tag} annotation scheme is followed by \citet{soares2019matching} and \citet{wu2019enriching} for span identification. In \citet{wu2019enriching}, each relation type $r \in \mathcal{R}$ is further expanded to two sub-classes as $r(e_1,e_2)$ and $r(e_2,e_1)$ to capture direction, while holding the \textit{s-tag} annotation as fixed. For DS-based RE, since the ordered tuple $(h,t)$ is given, the task is reduced to relation classification without direction. This side information is encoded in data with \textit{k-tag}, covering (2) but also (3) and (4). To account for (1), we also experiment with selective attention \citep{lin2016neural} which has been widely used in other works \citep{luo2017learning,han2018neural,alt2019fine}.  

\label{sec:3.2}

\begin{figure}[!t]
    \centering
    \includegraphics[width=0.45\textwidth]{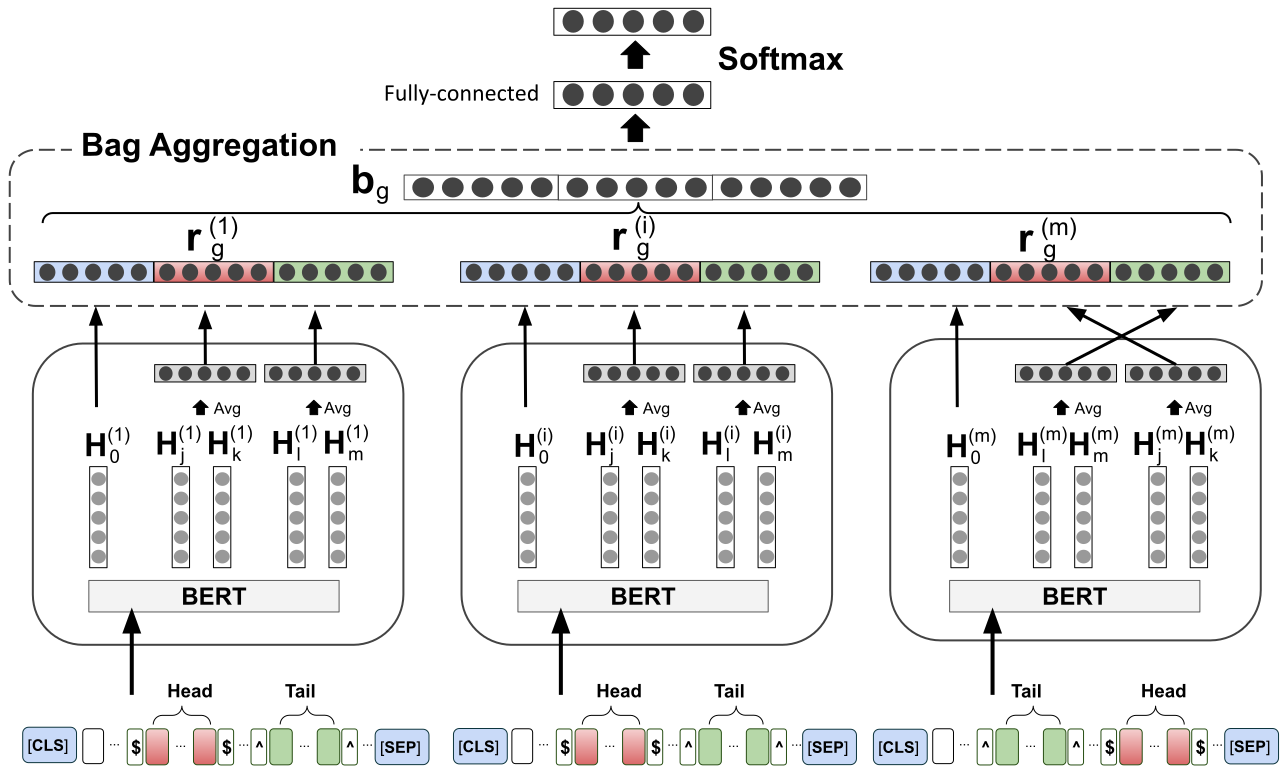}
    \caption{Multiple instance learning (MIL) based \textit{bag-level} relation classification BERT with KB ordered entity marking (Section \ref{sec:3.2}). Special markers \$ and \string^ always delimit the span of \textit{head} $(h_s,h_e)$ and \textit{tail} $(t_s,t_e)$ entities regardless of their order in the sentence. The markers captures the \textit{positions} of entities and latent relation \textit{direction}.}
    \label{fig:architecture}
\end{figure}

\subsection{Model Architecture}

BERT \citep{devlin2019bert} is used as our base sentence encoder, specifically, BioBERT \citep{lee2020biobert}, and we extend R-BERT \citep{wu2019enriching} to \textit{bag-level} MIL. Figure \ref{fig:architecture} shows the model’s architecture with \textit{k-tag}. Consider a bag $\mathcal{B}_g$ of size $m$ for a group $g \in \mathcal{G}$ representing the ordered tuple $(h,t)$, with corresponding spans $S_g=[(s_h^{(1)},s_t^{(1)}), ..., (s_h^{(m)},s_t^{(m)})]$ obtained with \textit{k-tag}, then for a pair of sentences in the bag and spans, $(s^{(i)}, (s_h^{(i)},s_t^{(i)}))$, we can represent the model in three steps, such that the first two steps represent the map $f$ and the final step $o$, as follows: \\

\noindent \textsc{1. Sentence Encoding}: BERT is applied to the sentence and the final hidden state $\mathbf{H}_0^{(i)} \in \mathbb{R}^d$, corresponding to the \texttt{[CLS]} token, is passed through a linear layer\footnote{Each linear layer is implicitly assumed with a bias vector} $\mathbf{W}^{(1)} \in \mathbb{R}^{d \times d}$ with $tanh(.)$ activation to obtain the global sentence information in $\mathbf{h}_0^{(i)}$.

\noindent \textsc{2. Relation Representation}: For the head entity, represented by the span $s_h^{(i)}=(j,k)$ for $k>j$, we apply average pooling $\frac{1}{k-j+1}\sum_{n=j}^{k} \mathbf{H}_n^{(i)}$, and similarly for the tail entity with span $s_t^{(i)}=(l,m)$ for $m>l$, we get $\frac{1}{m-l+1}\sum_{n=l}^{m} \mathbf{H}_n^{(i)}$. The pooled representations are then passed through a shared linear layer $\mathbf{W}^{(2)} \in \mathbb{R}^{d \times d}$ with $tanh(.)$ activation to get $\mathbf{h}_h^{(i)}$ and $\mathbf{h}_t^{(i)}$. To get the final latent relation representation, we concatenate the pooled entities representation with \texttt{[CLS]} as $\mathbf{r}_g^{(i)}=[\mathbf{h}_0^{(i)}; \mathbf{h}_h^{(i)}; \mathbf{h}_t^{(i)}] \in \mathbb{R}^{3d}$.

\noindent \textsc{3. Bag Aggregation}: After applying the first two steps to each sentence in the bag, we obtain $[\mathbf{r}_g^{(1)}, ..., \mathbf{r}_g^{(m)}]$. With a final linear layer consisting of a relation matrix $\mathbf{M}_r \in \mathbb{R}^{|\mathcal{R}| \times 3d}$ and a bias vector $\mathbf{b}_r \in \mathbb{R}^{|\mathcal{R}|}$, we aggregate the bag information with $o$ in two ways: 

\noindent \textbf{Average}: The bag elements are averaged as: 
\begin{equation*}
    \mathbf{b}_g = \frac{1}{m}\sum_{i=1}^m \mathbf{r}_g^{(i)}    
\end{equation*}

\noindent \textbf{Selective attention} \citep{lin2016neural}: For a row $\mathbf{r}$ in $\mathbf{M}_r$ representing the relation $r \in \mathcal{R}$, we get the attention weights as:
\begin{gather*}
    \alpha_i = \frac{\exp(\mathbf{r}^T\mathbf{r}_g^{(i)})}{\sum_{j=1}^{m} \exp(\mathbf{r}^T\mathbf{r}_g^{(j)})} \\
    \mathbf{b}_g = \sum_{i=1}^m \alpha_i\mathbf{r}_g^{(i)}
\end{gather*}
\noindent Following $\mathbf{b}_g$, a \textit{softmax} classifier is applied to predict the probability $p(r|\mathbf{b}_g;\theta)$ of relation $r$ being a true relation with $\theta$ representing the model parameters, where we minimize the cross-entropy loss during training.

\section{Experiments}

\subsection{Data}

Similar to \citep{dai2019distantly}, UMLS\footnote{We use 2019 release: \texttt{umls-2019AB-full}} \citep{bodenreider2004unified} is used as our KB and MEDLINE abstracts\footnote{\url{https://www.nlm.nih.gov/bsd/medline.html}} as our text source. A data summary is shown in Table \ref{table:data_summary} (see Appendix \ref{sec:A} for details on the data creation pipeline). We approximate the same statistics as reported in \citet{dai2019distantly} for relations and entities, but it is important to note that the data does not contain the same samples. We divided triples into train, validation and test sets, and following \citep{weston2013connecting,dai2019distantly}, we make sure that there is no overlapping facts across the splits. Additionally, we add another constraint, i.e., there is no sentence-level overlap between the training and held-out sets. To perform groups negative sampling, for the collection of evidence sentences supporting \texttt{NA} relation type bags, we extend KGC open-world assumption to \textit{bag-level} MIL (see \ref{sec:A.3}). 20\% of the data is reserved for testing, and of the remaining 80\%, we use 10\% for validation and the rest for training.

\label{sec:4.1}

\begin{table}[!h]
    \centering
    \caption{Overall statistics of the data.}
    \resizebox{7.6cm}{!}{
    \begin{tabular}{lllll}
        \toprule
        Triples & Entities & Relations & Pos. Groups & Neg. Groups \\
        \midrule
        169,438 & 27,403 & 355 & 92,070 & 64,448 \\
        \bottomrule
    \end{tabular}
    }
    \vspace{-0.2cm}
     \label{table:data_summary}
 \end{table}

\begin{table*}[!t]
    \centering
    \caption{Relation extraction results for different model configurations and data splits.}
    \resizebox{14cm}{!}{
    \begin{tabular}{cccccccccc}
    \toprule 
    Model & Bag Agg. & AUC & F1 & P@100 & P@200 & P@300 & P@2k & P@4k & P@6k \\
    \midrule
    \citet{dai2019distantly} & - & - & - & - & - & - & .913 & .829 & .753 \\ 
    \midrule
    \multirow{2}{*}{\textit{s-tag}} & \textit{avg} & .359 & .468 & .791 & .704 & .649 & .504 & .487 & .481 \\
    & \textit{attn} & .122 & .225 & .587 & .563 & .547 & .476 & .441 & .418 \\
    \midrule
    \multirow{2}{*}{\textit{s-tag+exprels}} & \textit{avg} & .383 & .494 & .508 & .519 & .521 & .507 & .508 & .511 \\
    & \textit{attn} & .114 & .216 & .459 & .476 & .482 & .504 & .496 & .484 \\
    \midrule
    \multirow{2}{*}{\textit{k-tag}} & \textit{avg} & \textbf{.684} & \textbf{.649} & \textbf{.974} & \textbf{.983} & \textbf{.986} & \textbf{.983} & \textbf{.977} & \textbf{.969} \\
    & \textit{attn} & .314 & .376 & .967 & .941 & .925 & .857 & .814 & .772 \\
    \bottomrule
    \end{tabular}
    }
     \label{table:main_results}
\end{table*}

\subsection{Models and Evaluation}

We compare each tagging scheme, \textit{s-tag} and \textit{k-tag}, with average (\textit{avg}) and selective attention (\textit{attn}) bag aggregation functions. To test the setup of \citet{wu2019enriching}, which follows \textit{s-tag}, we expand each relation type (\textit{exprels}) $r \in \mathcal{R}$ to two sub-classes $r(e_1, e_2)$ and $r(e_2, e_1)$ indicating relation direction from first entity to second and vice versa. For all experiments, we used batch size 2, bag size 16 with sampling (see \ref{sec:A.4} for details on bag composition), learning rate 2$e^{-5}$ with linear decay, and 3 epochs. As the standard practice \citep{weston2013connecting}, evaluation is performed through constructing candidate triples by combining the entity pairs in the test set with all relations (except \texttt{NA}) and ranking the resulting triples. The extracted triples are matched against the test triples and the precision-recall (PR) curve, area under the PR curve (AUC), F1 measure, and Precision@$k$, for $k$ in \{100, 200, 300, 2000, 4000, 6000\} are reported. 

\subsection{Results}

\begin{figure}[!t]
    \includegraphics[width=1.0\linewidth]{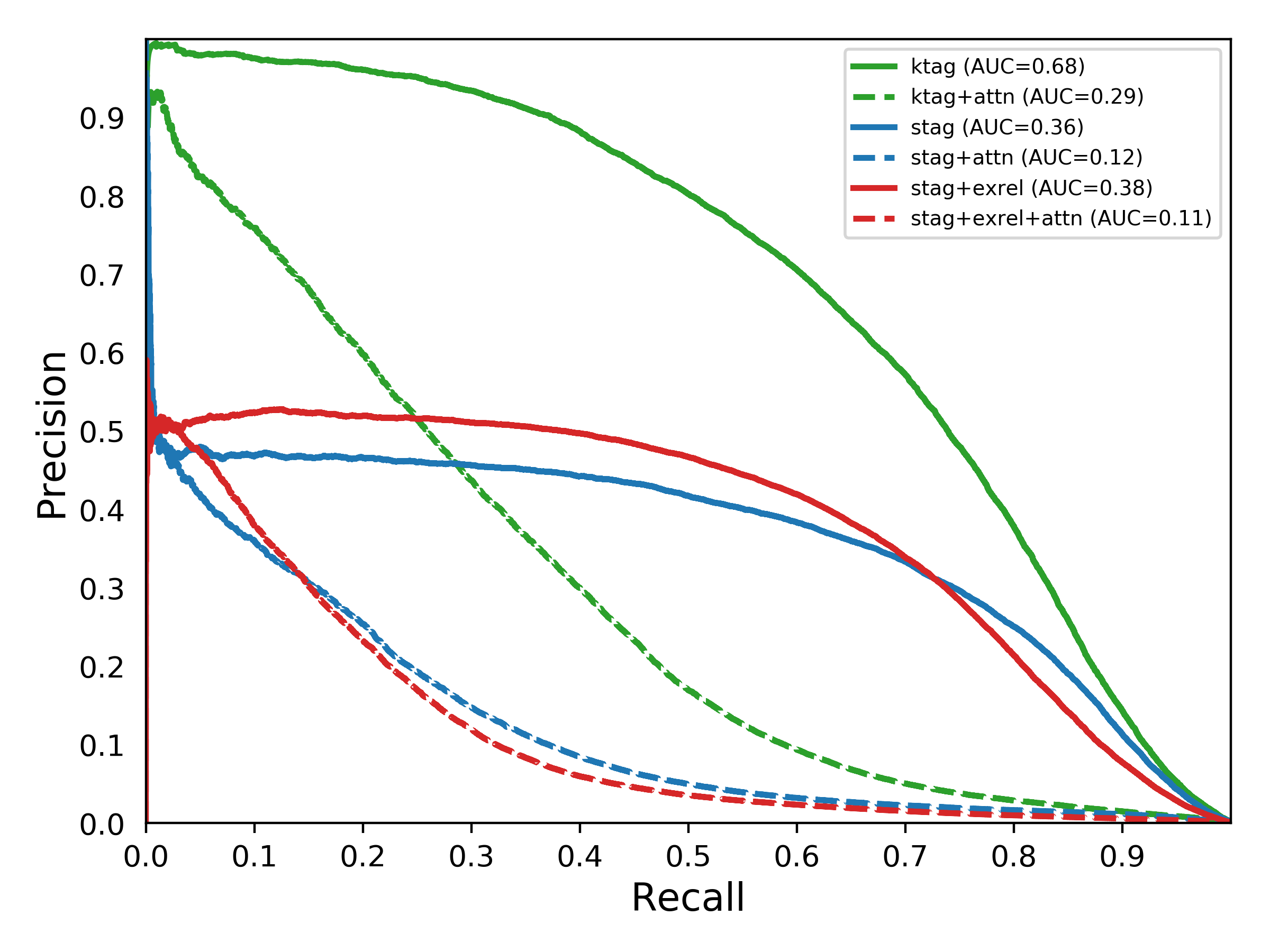}
    \centering
    \vspace{-0.5cm}
    \caption{Precision-Recall (PR) curve for different models. We see that the models with \textit{k-tag} perform better than the \textit{s-tag} with average aggregation showing consistent performance for long-tail relations.}
    \label{fig:pr_curve}
\end{figure}

Performance metrics are shown in Table \ref{table:main_results} and plots of the resulting PR curves in Figure \ref{fig:pr_curve}. Since our data differs from \citet{dai2019distantly}, the AUC cannot be directly compared. However, Precision@$k$ indicates the general performance of extracting the true triples, and can therefore be compared. Generally, models annotated with \textit{k-tag} perform significantly better than other models, with \textit{k-tag+avg} achieving state-of-the-art Precision@\{2k,4k,6k\} compared to the previous best \citep{dai2019distantly}. The best model of \citet{dai2019distantly} uses PCNN sentence encoder, with additional tasks of SimplE \citep{kazemi2018simple} based KGC and KG-attention, entity-type classification and named entity recognition. In contrast our data-driven method, \textit{k-tag}, greatly simplifies this by directly encoding the KB information, i.e., order of the \textit{head} and \textit{tail} entities and therefore, the latent relation direction. Consider again the example in Figure \ref{fig:overview} where our source triple $(h,r,t)$ is (\textit{neurofibromatosis 1}, \textit{associated\_genetic\_condition}, \textit{breast cancer}), and only last sentence has the same order of entities as KB. This discrepancy is conveniently resolved (note in Figure \ref{fig:architecture}, for last sentence the extracted entities sentence order is flipped to KG order when concatenating, unlike \textit{s-tag}) with \textit{k-tag}. We remark that such knowledge can be seen as learned, when jointly modeling with KGC, however, considering the task of \textit{bag-level} distant RE only, the KG triples are \textit{known} information and we utilize this information explicitly with \textit{k-tag} encoding.

As PCNN \citep{zeng2015distant} can account for the relative positions of head and tail entities, it also performs better than the models tagged with \textit{s-tag} using sentence order. Similar to \citet{alt2019fine}\footnote{Their model does not use any entity marking strategy.}, we also note that the pre-trained contextualized models result in sustained long tail performance. \textit{s-tag+exprels} reflects the direct application of \citet{wu2019enriching} to \textit{bag-level} MIL for distant RE. In this case, the relations are explicitly extended to model entity direction appearing first to second in the sentence, and vice versa. This implicitly introduces independence between the two sub-classes of the same relation, limiting the gain from shared knowledge. Likewise, with such expanded relations, class imbalance is further enhanced to more fine-grained classes. 

Though selective attention \citep{lin2016neural} has been shown to improve the performance of distant RE \citep{luo2017learning,han2018neural,alt2019fine}, models in our experiments with such an attention mechanism significantly underperformed, in each case bumping the area under the PR curve and making it flatter. We note that more than 50\% of bags are under-sized, in many cases, with only 1-2 sentences, requiring repeated over-sampling to match fixed bag size, therefore, making it difficult for attention to learn a distribution over the bag with repetitions, and further adding noise. For such cases, the distribution should ideally be close to uniform, as is the case with averaging, resulting in better performance. 

\section{Conclusion}

This work extends BERT to \textit{bag-level} MIL and introduces a simple data-driven strategy to reduce the noise in distantly supervised biomedical RE. We note that the \textit{position} of entities in sentence and the \textit{order} in KB encodes the latent \textit{direction} of relation, which plays an important role for learning under such noise. With a relatively simple methodology, we show that this can sufficiently be achieved by reducing the need for additional tasks and highlighting the importance of data quality.

\section*{Acknowledgements}

The authors would like to thank the anonymous reviewers for helpful feedback. The work was partially funded by the European Union's Horizon 2020 research and innovation programme under grant agreement No. 777107 through the project Precise4Q and by the German Federal Ministry of Education and Research (BMBF) through the project DEEPLEE (01IW17001).

\bibliography{acl2020}
\bibliographystyle{acl_natbib}

\appendix

\setcounter{table}{0}
\renewcommand{\thetable}{A.\arabic{table}}

\section{Data Pipeline}
\label{sec:A}
In this section, we explain the steps taken to create the data for distantly-supervised (DS) biomedical relation extraction (RE). We highlight the importance of a data creation pipeline as the quality of data plays a key role in the downstream performance of our model. We note that a pipeline is likewise important for generating reproducible results, and contributes toward the possibility of having either a benchmark dataset or a repeatable set of rules.

\subsection{UMLS processing} 

The fact triples were obtained for English concepts, filtering for \texttt{RO} relation types only \citep{dai2019distantly}. We collected 9.9M (CUI\_head, relation\_text, CUI\_tail) triples, where CUI represents the concept unique identifier in UMLS.

\subsection{MEDLINE processing}

From 34.4M abstracts, we extracted 160.4M unique sentences. To perform fast and scalable search, we use the Trie data structure\footnote{\url{https://github.com/vi3k6i5/flashtext}} to index all the textual descriptions of UMLS entities. In obtaining a clean set of sentences, we set the minimum and maximum sentence character length to 32 and 256 respectively, and further considered only those sentences where matching entities are mentioned only once. The latter decision is to lower the noise that may come when only one instance of multiple occurrences is marked for a matched entity.  With these constraints, the data was reduced to 118.7M matching sentences. 

\subsection{Groups linking and negative sampling}

Recall the entity groups $\mathcal{G} = \mathcal{G}^{+} \cup \mathcal{G}^{-}$ (Section \ref{sec:3.1}). For training with \texttt{NA} relation class, we generate hard negative samples with an open-world assumption \citep{soares2019matching,pbg} suited to \textit{bag-level} multiple instance learning (MIL). From 9.9M triples, we removed the relation type and collected 9M CUI groups in the form of $(h,t)$. Since each CUI is linked to more than one textual form, all of the text combinations for two entities must be considered for a given pair, resulting in 531M textual groups $\mathcal{T}$ for the 586 relation types. 

Next, for each matched sentence, let $\mathcal{P}_s^2$ denote the size 2 permutations of entities present in the sentence, then $\mathcal{T} \cap \mathcal{P}_s^2$ return groups which are \textit{present in KB} and \textit{have matching evidence} (positive groups, $\mathcal{G}^{+}$). Simultaneously, with a probability of $\frac{1}{2}$, we remove the $h$ or $t$ entity from this group and replace it with a novel entity $e$ in the sentence, such that the resulting group $(e,t)$ or $(h,e)$ belongs to $\mathcal{G}^{-}$. This method results in sentences that are seen both for the true triple, as well as for the invalid ones. Further using the constraints that the relation group sizes must be between 10 to 1500, we find 354\footnote{355 including \texttt{NA} relation} relation types (approximately the same as \citet{dai2019distantly}) with 92K positive groups and 2.1M negative groups, which were reduced to 64K by considering a random subset of 70\% of the positive groups. Table \ref{table:data_summary} provides these summary statistics.
\label{sec:A.3}

\subsection{Bag composition and data splits}

For bag composition, we created bags of constant size by randomly under- or over-sampling the sentences in the bag \citep{han-etal-2019-opennre} to avoid larger bias towards common entities \citep{soares2019matching}. The true distribution had a long tail, with more than 50\% of the bags having 1 or 2 sentences. We defined a bag to be \textit{uniform}, if the special markers represent the same entity in each sentence, either $h$ or $t$. If the special markers can take on both $h$ or $t$, we consider that bag to have a \textit{mix} composition. The \textit{k-tag} scheme, on the other hand, naturally generates uniform bags. Further, to support the setting of \citet{wu2019enriching}, we followed the \textit{s-tag} scheme and expanded the relations by adding a suffix to denote the directions as $r(e_1,e_2)$ or $r(e_2,e_1)$, with the exception of the \texttt{NA} class, resulting in 709 classes. For fair comparisons with \textit{k-tag}, we generated uniform bags with \textit{s-tag} as well, by keeping $e_1$ and $e_2$ the same per bag. Due to these bag composition and class expansion (in one setting, \textit{exprels}) differences, we generated three different splits, supporting each scheme, with the same test sets in cases where the classes are not expanded and a different test set when the classes are expanded. Table \ref{table:data_splits} shows the statistics for these splits.

\begin{table}[!h]
    \centering
    \caption{Different data splits.}
    \resizebox{7.7cm}{!}{
    \begin{tabular}{cccccc}
    \toprule 
    Model & Set Type & Triples & Triples (w/o NA) & Groups & Sentences (Sampled) \\
    \midrule
    \multirow{3}{*}{\textit{k-tag}} & train & 92,972 & 48,563 & 92,972 & 1,487,552 \\
                           & valid & 13,555 & 8,399 & 15,963 & 255,408 \\
                           & test & 33,888 & 20,988 & 38,860 & 621,760 \\
    \midrule
    \multirow{3}{*}{\textit{s-tag}} & train & 91,555 & 47,588 & 125,852 & 2,013,632 \\
                           & valid & 13,555 & 8,399 & 22,497 & 359,952 \\
                           & test & 33,888 & 20,988 & 55,080 & 881,280 \\
    \midrule
    \multirow{3}{*}{\textit{s-tag+exprels}} & train & 125,155 & 71,402 & 125,439 & 2,007,024 \\
                                   & valid & 22,604 & 16,298 & 22,607 & 361,712 \\
                                   & test & 55,083 & 39,282 & 55,094 & 881,504 \\
    \bottomrule
    \end{tabular}
    }
     \label{table:data_splits}
\end{table}

\label{sec:A.4}

\end{document}